\documentclass[runningheads]{llncs}
\usepackage[T1]{fontenc}
\usepackage{graphicx}
\usepackage{booktabs}
\usepackage[misc]{ifsym}
\newcommand{\corr}{(\Letter)}
\usepackage{xcolor}
\usepackage{multirow}
\usepackage{amsmath}
\usepackage{bbm}
\usepackage{array}

\begin{document}

\title{Video-Based Prediction of In-Flight Particle Characteristics in Atmospheric Plasma Spraying}

\titlerunning{Video-Based Prediction of In-Flight Particle Characteristics in APS}
% If the full title of your paper is short enough to also fit in the running head, you can omit the abbreviated paper title here. You can check as follows: if you comment out the \titlerunning line, something will appear in the header of all odd-numbered pages of your PDF from page 3 onward. This something is either the full title (in which case all is well), or the error message "Title Suppressed Due to Excessive Length". If this error message appears, you're going to want to provide an abbreviated title within the \titlerunning command, because if you won't do it, Springer will do it for you.

%N.B.: Author information (both in the \author{} and \authorrunning{} command) should only be present in the Camera-Ready Version of your paper. The version that you initially submit for review, ought to be double-blind. So, when initially submitting your paper, use:

%\author{Author information scrubbed for double-blind reviewing}
\author{
Abhijeet Praveen\inst{1,2} \and
Sareh Soleimani\inst{1,2} \and
Cormac Cureton\inst{1,2} \and
Aman Sidhu\inst{1,2} \and
Kintak Raymond Yu\inst{3} \and
Cristian Cojocaru\inst{3} \and
Narges Armanfard\inst{1,2}\corr
}
% You may leave out the orcidID information, if you want to.
% Use \corr to indicate the corresponding author. Note the spacing around the \corr command. Only one author can be the corresponding author.

%N.B.: comment out the \authorrunning{} command for the double-blind version of your paper submitted for review. Later, if your paper is accepted, use the command for the Camera-Ready Version.
\authorrunning{A. Praveen et al.}
% If there are two authors, write 'A.L. Benjamin and C.C. Broadus Jr.'
% If there are more than two authors, '[...] et al.' is used.

\toctitle{Video-Based Prediction of In-Flight Particle Characteristics in Atmospheric Plasma Spraying}
\tocauthor{Abhijeet Praveen, Sareh Soleimani, Cormac Cureton, Aman Sidhu, Kintak Raymond Yu, Cristian Cojocaru, Narges Armanfard}

\institute{
Department of Electrical and Computer Engineering, 
McGill University, Montreal, QC H2X 2G6, Canada\\
\email{\{abhijeet.praveen, cormac.cureton, aman.sidhu\}@mail.mcgill.ca} \\
\email{\{sareh.soleimanigilakjani, narges.armanfard\}@mcgill.ca}
\and
Mila -- Quebec AI Institute, Montreal, QC H2S 3H1, Canada
\and
National Research Council of Canada, Boucherville, QC J4B 6G4, Canada\\
\email{\{kintakraymond.yu, cristian.cojocaru\}@cnrc-nrc.gc.ca}
}
%\institute{Affiliations scrubbed for double-blind reviewing}

\maketitle              % typeset the header of the contribution

\begin{abstract}
Atmospheric plasma spraying (APS) is a widely used coating process in which in-flight particle temperature and velocity strongly influence coating quality. However, these particle characteristics are difficult to monitor continuously during operation, motivating the development of non-invasive data-driven diagnostic methods. In this work, we investigate the predictive potential of high-speed video observations of the plasma plume for estimating in-flight particle characteristics in APS. We introduce three different video-derived feature representations and evaluate them using Tabular Prior-Data Fitted Networks (TabPFN), convolutional neural networks (CNN), and classical regression baselines including Random Forest, Gradient Boosting, Support Vector Regression, and XGBoost. Experiments are conducted using grouped leave-one-out cross-validation on 126 labeled pre- and post-spray video recordings from 63 APS spray runs. Across the engineered feature experiments, TabPFN achieves the most consistent performance for temperature prediction, reaching $R^2=0.86$ using the combined feature representation. CNN models particularly perform stronger for velocity prediction, achieving $R^2$ of $0.81$. In addition, we evaluate models operating directly on raw video frames using pretrained CNNs and find that the highest performance is achieved by a pretrained CNN with a regression head with $R^2$ of $0.90$ and $0.82$ for temperature and velocity, respectively. The results demonstrate that video-derived plume information provides a promising and scalable foundation for non-invasive APS diagnostics and real-time process monitoring.

\keywords{Atmospheric Plasma Spraying \and Video-Based Diagnostics \and TabPFN \and Industrial Process Monitoring \and Video Feature Extraction}

\end{abstract}

\section{Introduction}
%%%%%%%%%%%%%%%%%%%%%%%%%%%%
%Please note that the first paragraph of a section or subsection is not indented. The first paragraph that follows a table, figure, equation etc. does not need an indent, either.
%Subsequent paragraphs, however, are indented.
%%%%%%%%%%%%%%%%%%%%%%%%%%%%%

% outline what APS is and what it is used for

% to consistently create high quality coatings, we must be able to model and capture the complex, non-linear effects of the input process parameters to the spray particles. past work has shown that we need to track temperature and velocity of these particles.

% emphasize electrode wear as an input feature given that prior methods might fail despite optimal parameters due to external factors which still influenece spray despite being not a direct input

% paragraph outlining gap in literature for video+process parameters+electrode wear as input feature

% paragraph for novelty

% short overview of paper

Atmospheric plasma spray (APS) is a popular thermal spray technique for enhancing surface material properties through application-specific coatings \cite{b2}. APS utilizes a plasma torch to generate a high-temperature, high-velocity plasma jet that melts and propels feedstock powder particles onto a substrate, forming the coating upon impact \cite{b3,singh2007use}.

However, consistently creating high-quality coatings requires precise control during the spray process \cite{b2}. Specifically, the properties of the final coating are directly influenced by the in-flight characteristics such as temperature and velocity of the particles as they are accelerated by the spray plume prior to impact \cite{b3}. Particles that are insufficiently heated or improperly accelerated result in poor adhesion and high porosity \cite{singh2007use}. Therefore, ensuring that particle temperature and velocity remain within a desired operating range is considered critical for reducing variability in coating properties caused by process variations \cite{Add}. The relationship between input spray process parameters, such as gas flow rates, electrical power, and standoff distance, and the resulting in-flight particle characteristics is inherently non-linear and sensitive to spray operating conditions, thereby requiring substantial expert knowledge to maintain consistent output quality \cite{b9,b10}. Compounding this challenge, the in-flight particle temperature and velocity are not directly observable during production, making real-time control and monitoring difficult for maintaining coating quality \cite{b3}. Although dedicated diagnostic systems such as DPV-2000 and Accuraspray have been widely used to measure individual in-flight particle temperatures and velocities via optical methods~\cite{b14}, these setups are typically expensive and are not designed for dynamic, real-time monitoring during active spraying~\cite{b11}.

Beyond the spray process parameters, recent work has demonstrated that factors external to the spray inputs can have a significant influence on coating quality. In particular, electrode usage, referring to the gradual erosion of the cathode and anode within the plasma torch over its operational lifetime, has been shown to alter the characteristics of the plasma jet and consequently the in-flight particle properties, even when all commanded process parameters remain unchanged \cite{b9,b10}. This means that a spray system operating under optimal parameter settings may still produce poor coatings as electrodes age, thereby introducing a form of process drift that is not captured by parameter-based predictive models. Therefore, models that can capture these process variations and reflect the evolving physical state of the spray process, beyond the commanded inputs are critical for robust prediction of APS particle characteristics.

Despite growing interest in data-driven approaches to APS process modeling, a gap remains in the literature with respect to the use of spray video recordings for predicting in-flight particle characteristics. Existing machine learning methods for predicting in-flight particle characteristics have largely relied on scalar process parameters alone, and while some have incorporated independent plume images for in-process fault detection~\cite{bokade2025thermal} or spray target classification~\cite{b11}, none have exploited the rich spatial and temporal information available from video observations of the spray plume. Non-invasive, camera-based diagnostics of the plasma jet and spray plume offer a complementary source of process state information, capturing phenomena such as plume shape, luminosity distribution, and jet instabilities, that are not reflected in process parameters alone~\cite{Add1}. Therefore, the integration of video-derived features represents an underexplored but promising direction for improving the accuracy and robustness of real-time prediction and monitoring of particle characteristics in APS environments.

%The integration of video-derived features alongside process parameters and electrode condition data therefore represents an underexplored but promising direction for improving the accuracy and robustness of real-time particle characteristic prediction in APS.

This work investigates the use of video observations of the plasma spray plume to predict in-flight particle characteristics, specifically particle temperature and velocity in APS. To leverage the information contained in plume observations, we first derive multiple feature representations from recorded videos. These include spectro-temporal descriptors that capture temporal intensity dynamics, plume motion, and boundary variability, as well as geometric descriptors that characterize the spatial structure and morphology of the plasma plume. These descriptors are computed either over temporal windows of video frames or averaged across the entire video segment.

We conduct the first phase of experiments using these engineered feature representations, where several machine learning models are evaluated for predicting particle characteristics. In particular, we investigate the use of Tabular Prior-Data Fitted Networks (TabPFN), a transformer-based model trained on synthetic tabular tasks that enables high-capacity tabular prediction in low-data regimes through in-context learning (ICL)~\cite{b29}. This property is particularly advantageous for APS applications, where experimental datasets are typically limited and costly to obtain. In addition, we evaluate several classical regression baselines including Random Forest \cite{breiman2001random}, Gradient Boosting \cite{friedman2001greedy}, Support Vector Regression (SVR) \cite{drucker1997support}, and XGBoost \cite{chen2016xgboost}. We also evaluate a lightweight Convolutional Neural Network (CNN) trained directly on the engineered feature representations.

In the second phase of experiments, we explore an alternative modeling strategy that directly leverages raw video observations of the spray plume. In this phase, visual features are extracted using a ResNet18 \cite{he2016deep} CNN pretrained on the ImageNet dataset \cite{deng2009imagenet}, which serves as a fixed feature extractor. Two architectures are evaluated, including a frozen pretrained CNN followed by a regression head, and a frozen pretrained CNN combined with a Long Short-Term Memory (LSTM) network to also capture temporal dependencies across video frames \cite{hochreiter1997long,donahue2015lrcn}. This comparison allows us to assess whether explicit temporal modeling of plume dynamics provides additional predictive benefit.

In summary, the primary contributions of this work are as follows:

\begin{enumerate}
\item The introduction of a video-based monitoring approach for APS particle diagnostics that leverages visual observations of the spray plume and enables the potential deployment of real-time monitoring models in industrial APS environments.
\item The design and evaluation of multiple video-derived feature representations, including spectro-temporal windowed features, geometric windowed features, and geometric averaged features, for capturing plume dynamics relevant to particle temperature and velocity.
\item A comprehensive comparison of several tabular and deep learning models for predicting particle characteristics across two experimental phases using engineered feature representations and raw video recordings.
\item Experimental results demonstrating that video-derived information provides strong predictive signals for particle characteristics, and that both engineered feature models and raw-video deep learning architectures can achieve strong predictive performance in the data-limited APS setting.
\end{enumerate}

The remainder of this paper is organized as follows. Section~\ref{sec:related-work} reviews prior research on APS diagnostics and machine learning approaches for predicting particle characteristics. Section~\ref{sec:methods} presents the methodological framework of this study, including the APS experimental setup and the video-derived feature representations used for modeling. Section~\ref{sec:experiments} describes the two experimental phases, including models applied to engineered feature representations and models operating directly on raw video frames, along with the evaluation metrics and the experimental results. Section~\ref{sec:discussion} interprets the experimental findings and discusses the implications for APS video-based monitoring. Finally, Section~\ref{sec:conclusion} presents the conclusions of this study and outlines directions for future research.

\section{Related Work}
\label{sec:related-work}

% emphasize need for datadriven methods in order to capture complex input/output relationship in spray processes
The use of machine learning (ML) and artificial intelligence (AI) to model and predict in-flight particle characteristics in APS has been an active area of research for over two decades ~\cite{b18,b5,choudhury2013extreme,zhu2020prediction,kanta2011intelligent,zhu2021application,guessasma2004analysis,b32,b35}. Early contributions established foundational frameworks for applying neural networks to thermal spray processes. For example, Guessasma \emph{et al.}~\cite{guessasma2004analysis} provided early empirical insight into the influence of APS parameters on adhesion properties, laying the groundwork for later machine learning approaches. This includes work by Kanta \emph{et al.} \cite{b18}, demonstrating that artificial neural networks (ANNs) could learn complex, nonlinear relationships between APS power parameters and in-flight particle characteristics such as temperature and velocity, enabling forward prediction of coating structural attributes. Recently, Bobzin \emph{et al.}~\cite{b5} presented a comprehensive ML-based framework targeting the prediction of in-flight particle properties directly from plasma spray process parameters, with the goal of constructing a digital twin of the thermal spray process. While their results reinforced prior findings that careful feature selection and high-quality training data are critical for reliable prediction, the motivation to their paper underpins a key problem with existing data-driven methods. Specifically, as noted in \cite{b24}, most models are trained on very small datasets, anywhere from fewer than 20 to at most 38 samples \cite{b25}, requiring particle spray prediction to operate in a data-limited regime. These limitations motivate the need for learning approaches that remain effective in data-limited regimes while exploiting richer sources of process information. 
%Recent advances in foundation models for tabular data, such as TabPFN, offer promising capabilities for learning from small datasets through in-context learning mechanisms.

Alongside ML modeling of the spray process, a complementary line of research has explored imaging and video analysis for monitoring and characterising thermal spray processes~\cite{Add1}, as studies have shown that high-speed imaging can provide valuable insights into particle plume dynamics~\cite{b11}. Bobzin \emph{et al.}~\cite{bobzin2021high} used high-speed videography to analyze plasma jet instability and fluctuations during the spray process, demonstrating that temporally resolved imaging contains rich process-state information.

Building on this direction, imaging has also been used for in-process fault detection in cold spray additive manufacturing. Bokade \emph{et al.}~\cite{bokade2025thermal} introduced ThermoAnoNet, a deep unsupervised learning framework that monitors substrate temperature profiles from thermal imaging to detect process anomalies such as increased deposition rates or nozzle clogging, demonstrating the potential of non-invasive imaging for real-time process monitoring.

More recently,~\cite{b11} proposed a multi-modal approach incorporating video features for spray plume classification using a pre-trained YOLO model. However, this work treats each observation independently and does not exploit the temporal structure of the data, thereby discarding contextual information that is critical for capturing the evolving dynamics of the spray plume. Moreover, their approach focuses on classification rather than regression-based prediction of particle characteristics.

%In contrast to prior studies that rely primarily on scalar process parameters or treat visual observations only for static plume characterization, this work investigates video-based prediction of in-flight particle temperature and velocity using both engineered feature representations and learned temporal models. Specifically, we introduce and evaluate three complementary video-derived feature sets, namely spectro-temporal windowed (STW), geometric windowed (GW), and geometric averaged (GA) representations, designed to capture both dynamic and structural properties of the spray plume. We further provide a systematic comparison between high-capacity tabular learning with TabPFN, deep learning approaches based on CNN and CNN+LSTM architectures, and classical regression baselines including Random Forest, Gradient Boosting, SVR, and XGBoost. Through grouped leave-one-out evaluation in a data-limited setting, our study provides, to the best of our knowledge, the first comprehensive benchmark of temporally informed video-based APS particle prediction across both engineered and end-to-end learned visual representations.

\section{Methodology}
\label{sec:methods}
This section describes the APS experimental setup, video pre-processing and feature extraction methods used for predicting in-flight particle characteristics.

%\subsection{APS Dataset}
%\label{sec:dataset}

%This section presents the APS experimental setup and data acquisition procedure used in this study. The goal of the experiments was to record plasma plume imagery together with corresponding measurements of in-flight particle temperature and velocity in order to support data-driven modelling of particle characteristics.

\subsection{APS Experimental Setup}

All experiments were conducted using an Axial3Plus APS torch (Northwest Mettech Corp., Vancouver, BC, Canada). The system was configured with three pairs of new electrodes and operated with a powder feeder speed of 1 RPM. Experiments were performed using three nozzle diameters (3/8", 1/2", and 5/16"), while the torch current was maintained at a constant value of 125 A.

Particle diagnostic measurements were obtained using an Accuraspray 4.0 system (Tecnar, St-Bruno, QC, Canada), which provides optical measurements of in-flight particle temperature and velocity. Video recordings of the plasma plume were captured using a fixed RGB scientific camera.

Each experiment followed an identical spray procedure. After plasma ignition and stabilization, powder injection was initiated and a stationary diagnostic measurement was recorded prior to deposition (pre-spray). The torch then executed five spray passes across the substrate to deposit the coating. Upon completion of the deposition process, the torch returned to a stationary position where a second diagnostic measurement was collected (post-spray). 

Particle temperature and velocity labels were obtained separately for the pre-spray and post-spray intervals using the mean value of the diagnostic measurements within each interval. Corresponding video recordings were captured during both diagnostic phases, allowing each video to be associated with a particle characteristic label.

Five spray configurations with different gas flow rates were used as summarized in Table~\ref{tab:spray_conditions}. In total, 63 labeled spray runs were collected, resulting in 126 video recordings (63 pre-spray and 63 post-spray), each with corresponding diagnostic measurements and video data.

\begin{table}[ht]
\centering
\caption{Spray Conditions Used for Data Collection (Gas Flow Rates in LPM)}
\label{tab:spray_conditions}
\begin{tabular}{c c c c}
\toprule
\textbf{Condition \#} & \textbf{Argon} & \textbf{Nitrogen} & \textbf{Hydrogen} \\
\midrule
1 & 125 & 75 & 50 \\
2 & 175 & 50 & 25 \\
3 & 200 & 25 & 25 \\
4 & 187 & 37.5 & 25 \\
5 & 187 & 25 & 37.5 \\
\bottomrule
\end{tabular}
\end{table}

All experiments were implemented in Python and executed using an NVIDIA RTX 5000 Ada Generation GPU (32 GB, CUDA 12.6).

%\subsection{Video Pre-processing}
%\label{sec:video_acquisition}

%The camera was positioned to observe the active torch plume region during operation. The resulting videos provide direct visual access to plume behaviour, including changes in luminosity, plume extent, spatial structure, and short-term fluctuations in the spray jet. Such visual cues are potentially informative because they reflect changes in the plasma state and particle transport conditions, both of which influence the resulting in-flight particle temperature and velocity.

\begin{figure}[t]
    \centering
    \includegraphics[width=\textwidth]{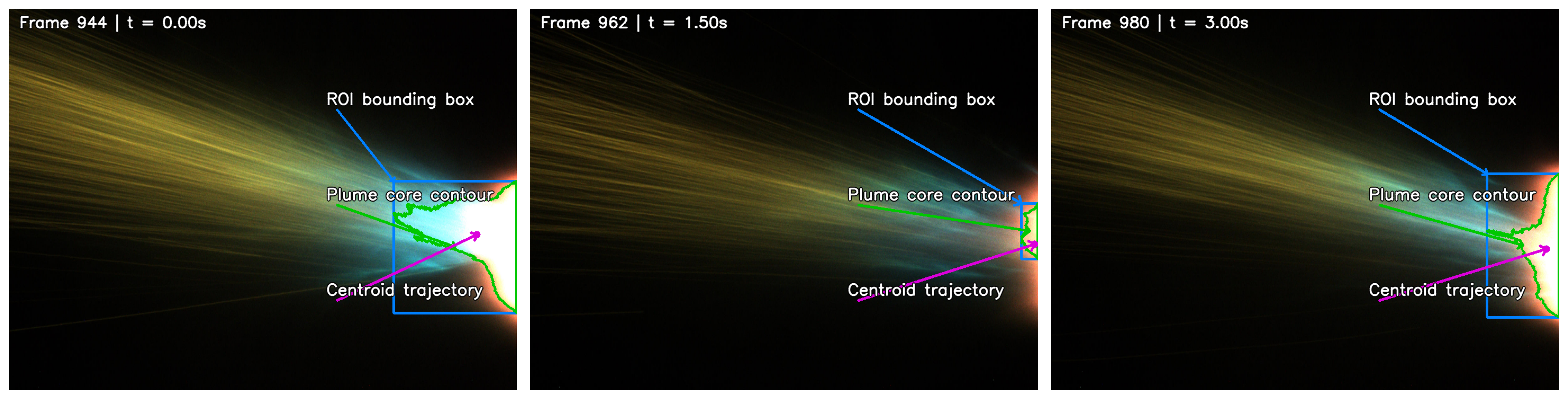}
    \caption{Representative first, second and third temporal frames from APS plume videos (from left to right), illustrating the visual information available in the spray recordings. The overlays show the segmented plume core, the region of interest (ROI) used for feature extraction, and the centroid trajectory used to characterize plume motion. These visual cues form the basis for the engineered video feature representations described in Section~\ref{sec:video_features}.}
    \label{fig:video_frames}
\end{figure}

\subsection{Video Feature Representations}
\label{sec:video_features}

\paragraph{Pre-processing.} To ensure consistency across experiments, each video was first converted to grayscale and each recorded interval was temporally cropped to the active torch period. The videos were recorded at 15 frames per second (FPS) and each interval was processed as a fixed-duration six-second spray segment. Video durations varied slightly across experiments, therefore the segment length was selected based on the shortest usable interval in the dataset.
This ensured that all videos were represented using a consistent temporal footprint for feature extraction and learning. Frames were then extracted from the cropped video interval. To reduce redundancy and computational cost, videos were downsampled by keeping every $k$-th frame, where $k=5$ in our implementation. Feature extraction was then performed either over the entire video segment or over a set of fixed temporal windows. For the windowed representations, each video was divided into three equal-length windows, of two seconds each, in the downsampled frame domain. 
Figure~\ref{fig:video_frames} shows representative frames from the three temporal windows of a sample video. These frames illustrate temporal variations in plume size and shape, motivating the use of representations that capture both global plume geometry and short-term temporal dynamics.
 
To capture these different aspects of plume behavior, we construct three complementary video-derived feature representations: Geometric Averaged (GA), Geometric Windowed (GW), and Spectro-Temporal Windowed (STW).

\paragraph{Notation.} Let a pre-processed video segment be denoted $V=\{I_1,\dots,I_T\}$, where $T$ is the number of frames in the segment and $I_t \in [0,255]^{H\times W}$ is the $t$-th grayscale frame, restricted to the plume region of interest (ROI) of height $H$ and width $W$ pixels. We write $p=(r,c)$ for a pixel at row $r$ and column $c$, and $I_t(p)$ for its intensity. A binary plume mask $M_t \in \{0,1\}^{H\times W}$ is obtained by global intensity thresholding, $M_t(p) = \mathbbm{1}\!\left(I_t(p) \ge \tau\right)$, with a fixed threshold $\tau=200$ that isolates the bright plume core, where $\mathbbm{1}(\cdot)$ is the indicator function. Plume contours are extracted as the external boundaries of the connected components of $M_t$, and the plume centroid $c_t \in \mathbbm{R}^2$ is the mean position of all mask pixels in frame $t$.

\paragraph{Geometric Averaged (GA).}
The GA representation captures global plume morphology using nine frame-level descriptors. First, intensity-distribution features are computed over five pixel-intensity bins $B_k$ with bin edges $\{0,50,100,150,200,255\}$, as the fraction of pixels falling in each bin pooled over all frames of the segment:
\begin{equation}
    b_k = \frac{1}{H W T}\sum_{t=1}^{T}\sum_{p}
    \mathbbm{1}\!\left(I_t(p)\in B_k\right),
    \qquad k=1,\dots,5,
    \label{eq:bins}
\end{equation}
where $HWT$ is the total number of pixels across all $T$ frames, so that each $b_k$ is the fraction of pixels falling in intensity bin $B_k$, providing a coarse description of the plume's radiative structure. Second, four geometric descriptors are computed per frame from the plume mask and averaged over the segment: the per-frame plume area $A_t$ (the
total number of mask pixels enclosed by the external contours of $M_t$), the contour perimeter $P_t$ (the total length of those contours, in pixels), and the vertical and horizontal extents $H_t$ and $W_t$, defined as the number of image rows (respectively columns) whose count of mask pixels exceeds 10\% of the maximum row (column) count, i.e., $H_t = \bigl|\{\,r : \textstyle\sum_{c} M_t(r,c) \ge 0.1 \max_{r'} \sum_{c} M_t(r',c)\,\}\bigr|$, where $r'$ indexes over all rows, and
analogously for $W_t$. This yields a compact 9-dimensional representation per video.

\paragraph{Geometric Windowed (GW).}
While GA summarizes the plume over the full video, GW preserves temporal variation by computing the same nine descriptors separately within each of the three temporal windows $\mathcal{W}_1,\mathcal{W}_2,\mathcal{W}_3$
and concatenating them across time, resulting in $9\times3=27$ features per video.

\paragraph{Spectro-Temporal Windowed (STW).}
The STW representation describes short-term temporal dynamics of the plume. For these features, frames are first contrast-enhanced using Contrast Limited Adaptive Histogram Equalization (CLAHE)~\cite{zuiderveld1994clahe}, and the per-frame centroid $c_t$ and perimeter $P_t$ are computed from the largest external contour of the plume mask. Within each temporal window $\mathcal{W}_w$ ($w\in\{1,2,3\}$) of $T_w$ frames, let $s(t)$ denote the mean ROI intensity of frame $t$, and let $S(f) = \sum_{t} \tilde{s}(t)\, e^{-\imath 2\pi f t}$ be the discrete Fourier transform of the mean-removed, Hann-windowed signal $\tilde{s}(t)$ at frequency $f$ (in
Hz), so that $|S(f)|^2$ is its power spectrum. Eight descriptors are computed per window:
(i) the temporal dominant frequency $f^{\ast} = \arg\max_{f>0} |S(f)|^2$;
(ii) the band power in the 5--20\,Hz range,
$\mathrm{BP} = \sum_{5 \le f < 20} |S(f)|^2$,
capturing mid-frequency oscillatory energy;
(iii) the autocorrelation time $\tau_{\mathrm{ac}}$, the smallest lag at which the normalized autocorrelation of $s(t)$ first becomes non-positive, measuring temporal persistence;
(iv) the mean centroid speed, computed as the average frame-to-frame displacement of the plume centroid scaled by the frame rate, following standard centroid-tracking approaches~\cite{maggio2011video};
(v) the centroid along--cross ratio, the ratio of the mean absolute centroid displacement parallel to the spray axis to that perpendicular to it, $\rho = \overline{|\Delta c^{\parallel}|} \,/\,
\overline{|\Delta c^{\perp}|}$, where $\Delta c_t = c_{t+1}-c_t$ is the inter-frame centroid displacement, $\Delta c^{\parallel}$ and $\Delta c^{\perp}$ are its components parallel and perpendicular to the spray axis, and $\overline{(\cdot)}$ denotes averaging over the window,
quantifying whether plume motion is primarily aligned with the spray direction;
(vi) the standard deviation of frame-to-frame perimeter differences, $\sigma_{\Delta P} = \mathrm{std}\!\left(P_{t+1}-P_t\right)$, capturing short-term boundary variability;
(vii) the 90th percentile of detected streak lengths, where streaks are elongated bright structures segmented from unsharp-masked frames and measured by the major axis of their minimum-area bounding rectangles, used as a proxy for persistent elongated flow structures; and
(viii) the mean intensity $\bar{s}$ within the window.
Concatenating the eight descriptors across the three windows yields $8\times3=24$ features per video. These spectro-temporal descriptors are commonly used in dynamic texture and spatio-temporal video analysis to characterize time-varying visual phenomena \cite{Doretto2003,Jansson_2018}.

\begin{figure}[t]
    \centering
    \includegraphics[width=\textwidth]{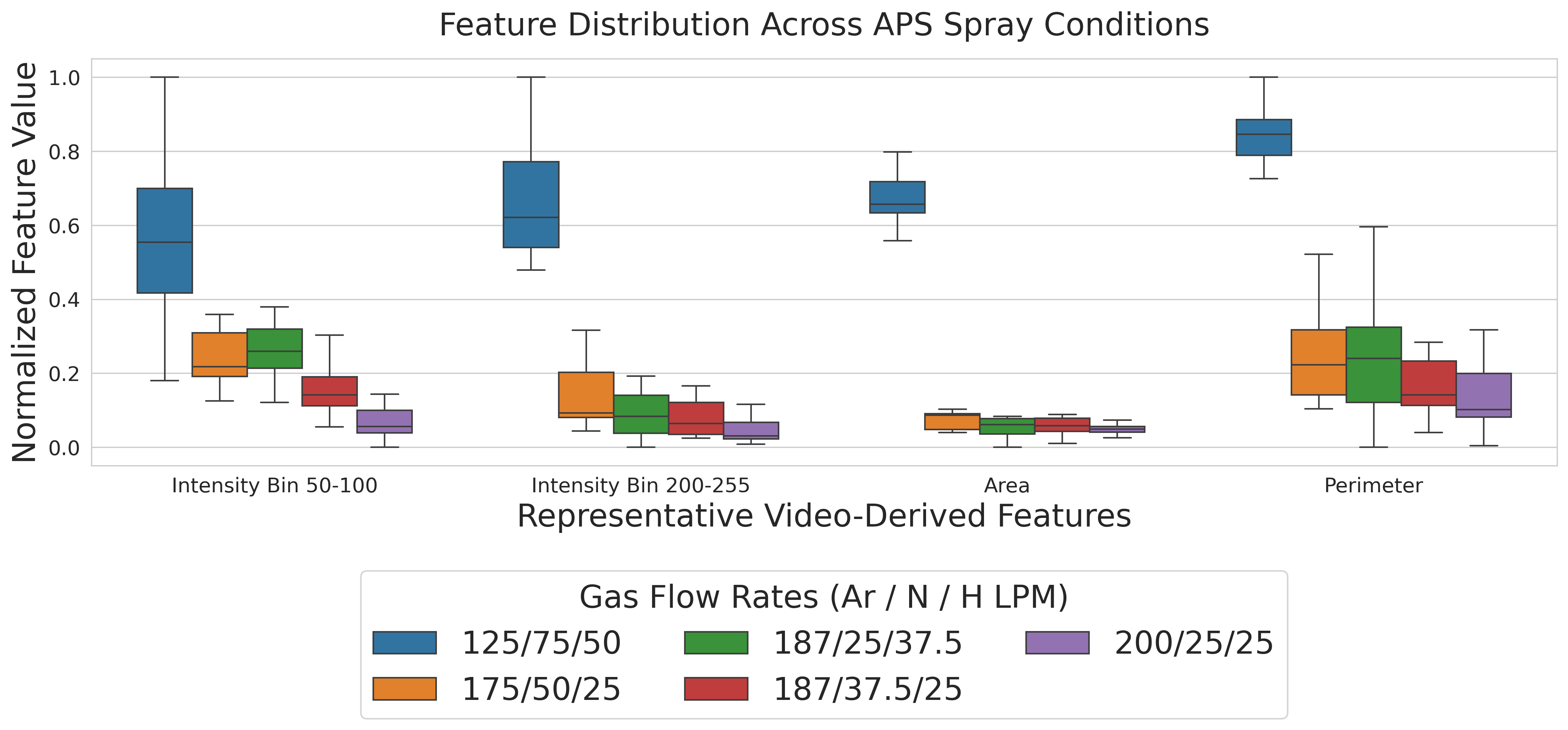}
    \caption{Distribution of representative video-derived plume features across APS spray conditions. Each grouped boxplot compares the normalized distributions of selected intensity-based and geometric descriptors under different gas flow configurations (Argon/Nitrogen/Hydrogen in LPM).}
    \label{fig:feature_distribution_grouped}
\end{figure}

The three feature sets capture complementary information. GA provides a stable global summary of plume structure, GW preserves time-localized geometric evolution, and STW focuses on dynamic temporal behavior that reflects transient flow instabilities and particle transport effects.
Figure~\ref{fig:feature_distribution_grouped} presents the distributions of several representative video-derived features across the five APS spray conditions. 
The selected descriptors include both intensity-based and geometric plume measurements, normalized to allow comparison across features. Consistent shifts in feature distributions are observed across different gas flow configurations, suggesting that the extracted plume descriptors capture condition-dependent variations in plume structure. These results provide qualitative evidence that video-derived features contain physically meaningful information related to APS operating parameters.

\section{Experiments}
\label{sec:experiments}

This section describes the modeling approaches used in this study and their evaluation. We first present the models applied to engineered feature representations, followed by models operating directly on raw video frames. We then evaluate these approaches to assess their effectiveness in predicting in-flight particle temperature and velocity.

\subsection{Models and Baselines for Engineered Feature Representations}
\label{sec:engineered_sets}

In the first phase of experiments, we evaluate both tabular machine learning models and deep learning approaches on the engineered video feature representations described in Section~\ref{sec:video_features}. The goal of this comparison is to assess whether the proposed plume descriptors can be effectively exploited by different learning paradigms in the data-limited APS setting. We consider several tabular regression models, including TabPFN and classical models such as Random Forest, Gradient Boosting, Support Vector Regression, and XGBoost. In addition, we evaluate a CNN applied to the engineered windowed feature representations.

\paragraph{TabPFN.} TabPFN is a transformer-based tabular model designed for small tabular datasets, where it has been shown to perform strongly through in-context learning rather than conventional task-specific training~\cite{b29,muller_transformers_2022}. This makes it particularly attractive in the APS setting, where labeled experiments are limited and collecting additional data is costly. Instead of learning solely through gradient updates on the target dataset, TabPFN leverages knowledge acquired from synthetic prior tasks and adapts to the target problem at inference time. In this study, separate single-task TabPFN regressors are used for particle temperature and particle velocity prediction.

\paragraph{CNN Architecture.} Motivated by prior work demonstrating the effectiveness of convolutional neural networks for monitoring thermal spray processes~\cite{b11}, we also implement a CNN architecture to model the engineered feature representations extracted from multiple temporal windows of video data. Each sample is represented as a tensor of shape $[B, F, T]$, where $B$ denotes the batch size, $F$ the number of engineered features, and $T=3$ the number of temporal windows corresponding to the first, second, and third temporal windows of the analyzed video. This representation allows the network to learn relationships between features across neighboring temporal windows. The CNN architecture consists of three stacked one-dimensional convolutional layers with 64 hidden channels. The first convolution uses a kernel size of 2 with padding 1, the second uses a kernel size of 3 with padding 1, and the third uses a kernel size of 2 without additional padding. Each convolutional layer is followed by a ReLU activation function. These convolutional filters operate along the temporal dimension and enable the network to capture short-range interactions and transitions between adjacent windows. After the convolutional layers, an adaptive average pooling layer aggregates the temporal dimension to produce a fixed-length representation. The resulting feature vector is then passed to a regression head composed of two fully connected layers with a ReLU activation between them, specifically $\text{Linear}(64, 64)$ followed by $\text{Linear}(64, 1)$, to produce the final scalar prediction corresponding to either particle temperature or particle velocity. Model hyperparameters were selected through validation experiments during model development, with the final configuration using the Adam optimizer, a learning rate of $1\times10^{-3}$, batch size 16, and 100 training epochs.

\paragraph{Classical Baselines.} For each baseline we use a single fixed, manually chosen configuration held constant across all feature sets and cross-validation folds, rather than tuning per fold. The Random Forest, Gradient Boosting, and SVR baselines are implemented using scikit-learn,
and XGBoost using its official library: Random Forest with 300 trees and unrestricted depth; Gradient Boosting with 300 estimators, learning rate 0.05, and maximum depth 3; SVR with an RBF kernel, $C{=}10$, and $\epsilon{=}0.1$; and XGBoost with 300 estimators, learning rate 0.05, maximum depth 4, and row and column subsampling of 0.9. Input features are
standardized to zero mean and unit variance using training-fold statistics only, with near-constant features removed and missing values imputed by training-fold means.

\subsection{Raw Video Modeling}
\label{sec:raw_features}
In the second phase of experiments, we also evaluate deep learning models that operate directly on raw video frames of the plasma plume. This phase aims to assess whether visual representations learned directly from plume imagery can capture predictive information related to particle temperature and velocity without relying on manually engineered descriptors.

For each experiment, the video segment corresponding to the measurement interval is extracted using the start and end timestamps of the active period provided in the dataset. To keep the experiment uniform, 24 frames were extracted from a fixed length of six seconds across all videos, as the full video lengths varied across all available samples. Each frame is resized to a fixed spatial resolution of $224\times224$ pixels and normalized before being passed to the neural network models.

To leverage existing visual representations while maintaining stability in the small-data APS setting, we employ a pretrained convolutional neural network as a fixed feature extractor. Specifically, we use ResNet18 \cite{he2016deep}, pretrained on the ImageNet dataset \cite{deng2009imagenet}, which processes each frame independently and produces a compact feature embedding representing the spatial structure and intensity patterns of the plasma plume.

Two architectures are evaluated using these extracted visual embeddings. The first architecture consists of a frozen pretrained CNN followed by a regression head. In this configuration, the CNN backbone remains fixed and only the final fully connected layers are trained to map the extracted visual features to the target particle characteristic. The regression head uses the same architecture as the CNN model described in Section~\ref{sec:engineered_sets}, consisting of two fully connected layers with ReLU activation between them ($\text{Linear}(64, 64)$ followed by $\text{Linear}(64, 1)$). This architecture evaluates whether frame-level spatial representations alone are sufficient for predicting particle temperature and velocity.

The second architecture extends this framework by introducing temporal modeling through a Long Short-Term Memory (LSTM) network. While the frozen CNN extracts spatial features from individual frames, plume dynamics in plasma spraying evolve continuously over time, and these temporal fluctuations may contain additional information related to particle temperature and velocity. To capture such sequential dependencies, each frame of a video sample is first processed by the frozen pretrained CNN to produce a sequence of frame-level embeddings. These embeddings are then passed to a single-layer LSTM network that models temporal relationships across the frame sequence. The final hidden state of the LSTM is used as the aggregated video representation and is subsequently passed through the same two-layer regression head  to generate the final prediction. LSTM networks are generally well suited for modeling sequential data and long-range temporal dependencies \cite{hochreiter1997long}, and CNN--LSTM architectures have been widely used for modeling temporal dynamics in video data \cite{donahue2015lrcn}.

%To provide a broader comparison, we also evaluate several classical regression baselines that are widely used in industrial prediction tasks. These include Random Forest regression \cite{breiman2001random}, Gradient Boosting regression \cite{friedman2001greedy}, SVR \cite{drucker1997support}, and Extreme Gradient Boosting (XGBoost) \cite{chen2016xgboost}. These methods represent strong nonlinear baselines for tabular prediction and allow us to assess whether the proposed video-derived representations provide predictive value beyond established machine learning approaches.

%For deep learning on engineered windowed features, we employ a convolutional neural network (CNN) that operates over the temporally ordered feature windows \cite{lecun1998gradient}. This architecture enables the model to learn local patterns across consecutive windows while preserving the temporal ordering of the extracted descriptors.
\subsection{Evaluation Metrics}
\label{sec:metrics}
All models are evaluated using grouped leave-one-out cross-validation (LOOCV). In each fold, all samples associated with one spray experiment identifier are held out for testing, while the remaining experiments are used for training. Therefore, each fold holds out both the pre-spray and post-spray videos from one spray run. This grouping is necessary because each spray experiment contributes multiple related video samples, corresponding to the pre-spray and post-spray intervals. Splitting these related samples across training and testing would introduce information leakage. Grouped LOOCV therefore provides a more realistic estimate of generalization to unseen spray runs while remaining appropriate for the limited dataset size.

Because each fold holds out only the two videos associated with a single spray run, fold-level performance estimates are not statistically meaningful. All reported metrics are therefore computed over the pooled out-of-fold predictions aggregated across all 63 folds, such that every
sample is evaluated exactly once as unseen test data. To quantify uncertainty, we additionally compute 95\% confidence intervals for $R^2$ via cluster bootstrap resampling over spray runs; these intervals are reported for the best-performing models in Section~\ref{sec:results}.

Model performance is reported separately for temperature and velocity prediction using the coefficient of determination ($R^2$) and the Percentage Accuracy Metric (PAM) as shown in \eqref{eq:r2} and \eqref{eq:pam}, respectively. For a set of $n$ true labels $y_i$ and corresponding predictions $\hat{y}_i$, $R^2$ is defined as:
\begin{equation}
    R^2=1-\frac{\sum_{i=1}^n(y_i-\hat y_i)^2}{\sum_{i=1}^n(y_i-\bar y)^2} \qquad \text{where } \bar y=\frac{1}{n}\sum_{i=1}^ny_i
    \label{eq:r2}
\end{equation}

The Percentage Accuracy Metric at tolerance $\delta\%$, denoted PAM@$\delta\%$, is defined as the proportion of predictions whose relative error is within $\delta\%$ of the ground truth:

\begin{equation}
    \text{PAM}@\delta\% = \frac{1}{n}\sum_{i=1}^{n}
    \mathbbm{1}
    \left(
        \left|\frac{\hat{y}_i-y_i}{y_i}\right| \le \frac{\delta}{100}
    \right)
    \label{eq:pam}
\end{equation}

Where $\mathbbm{1}(\cdot)$ is the indicator function, which equals 1 if the condition holds and 0 otherwise. We report PAM@2.5\%, PAM@5\%, and PAM@10\% in addition to $R^2$. While $R^2$ captures overall goodness of fit, PAM provides a more application-oriented measure of predictive reliability by quantifying how often predictions fall within practically meaningful tolerance bands. This is particularly useful in APS, where the temperature and velocity targets differ substantially in magnitude and where relative prediction error is often more informative than absolute error alone. In this sense, PAM@$\delta\%$ directly reflects diagnostic utility, measuring how reliably a model can confirm that particle temperature and velocity remain within a tolerance band around their target values, which is precisely the monitoring task performed in practice to keep the
process inside its operating window~\cite{Add}.

\subsection{Experimental Results and Analysis}
\label{sec:results}

The predictive performance of the proposed video-based APS modeling approaches under both experimental phases was evaluated using the $R^2$ and PAM metrics defined in Section~\ref{sec:metrics}.

\subsubsection{Performance Analysis}

\begin{table}[htbp]
\caption{Model performance comparison across all the engineered feature representations for particle temperature and velocity prediction.}
\label{tab:baseline_results}
\centering
\footnotesize
\setlength{\tabcolsep}{3pt}

\begin{tabular}{lcccccccc}
\toprule
\multirow{2}{*}{Model} & \multicolumn{4}{c}{Temperature Prediction} & \multicolumn{4}{c}{Velocity Prediction} \\
\cmidrule(lr){2-5} \cmidrule(lr){6-9}
 & $R^2$ & PAM$_{2.5}$ & PAM$_5$ & PAM$_{10}$ & $R^2$ & PAM$_{2.5}$ & PAM$_5$ & PAM$_{10}$ \\
\midrule

\multicolumn{9}{l}{\textbf{STW}} \\
Random Forest      & 0.51 & 0.23 & 0.40 & 0.63 & 0.07 & \underline{0.19} & 0.29 & \underline{0.57} \\
Gradient Boosting  & 0.43 & 0.24 & \underline{0.44} & \underline{0.66} & -0.04 & 0.14 & 0.29 & 0.49 \\
SVR                & -0.06 & 0.21 & 0.38 & 0.51 & \underline{0.12} & 0.12 & 0.23 & 0.44 \\
XGBoost            & 0.48 & 0.21 & 0.40 & 0.65 & 0.04 & 0.13 & 0.26 & 0.52 \\
TabPFN             & \underline{0.58} & \textbf{0.34} & \textbf{0.54} & \textbf{0.79} & 0.09 & 0.13 & \underline{0.33} & 0.52 \\
CNN & \textbf{0.71} & \underline{0.32} & \underline{0.44} & \underline{0.66} & \textbf{0.50} & \textbf{0.25} & \textbf{0.44} & \textbf{0.68} \\

\midrule
\multicolumn{9}{l}{\textbf{GW}} \\
Random Forest      & \textbf{0.86} & \underline{0.43} & 0.67 & \underline{0.89} & 0.36 & \underline{0.22} & 0.46 & 0.67 \\
Gradient Boosting  & \underline{0.85} & \underline{0.43} & 0.70 & \textbf{0.90} & 0.28 & \underline{0.22} & 0.44 & 0.68 \\
SVR                & 0.32 & 0.28 & 0.44 & 0.58 & 0.29 & 0.14 & 0.31 & 0.58 \\
XGBoost            & 0.82 & 0.40 & \underline{0.72} & 0.88 & 0.31 & 0.18 & 0.44 & 0.69 \\
TabPFN             & 0.83 & \textbf{0.53} & \textbf{0.75} & 0.88 & \underline{0.54} & \underline{0.22} & \underline{0.52} & \underline{0.74} \\
CNN & 0.73 & 0.30 & 0.45 & 0.68 & \textbf{0.81} & \textbf{0.48} & \textbf{0.67} & \textbf{0.85} \\

\midrule
\multicolumn{9}{l}{\textbf{GA}} \\
Random Forest      & \textbf{0.79} & \textbf{0.41} & \textbf{0.63} & \underline{0.79} & \underline{0.27} & \underline{0.17} & \underline{0.36} & 0.52 \\
Gradient Boosting  & 0.75 & 0.35 & 0.54 & 0.75 & 0.21 & 0.15 & 0.33 & \underline{0.55} \\
SVR                & 0.34 & 0.27 & 0.43 & 0.57 & 0.25 & 0.09 & 0.30 & 0.52 \\
XGBoost            & 0.72 & 0.29 & 0.51 & 0.75 & 0.16 & 0.10 & 0.26 & 0.52 \\
TabPFN             & \underline{0.78} & \underline{0.40} & \underline{0.59} & \textbf{0.81} & \textbf{0.45} & \textbf{0.22} & \textbf{0.41} & \textbf{0.61} \\

\midrule
\multicolumn{9}{l}{\textbf{STW+GW+GA}} \\
Random Forest      & \underline{0.80} & \underline{0.42} & \underline{0.71} & 0.85 & 0.52 & \underline{0.25} & 0.46 & 0.67 \\
Gradient Boosting  & 0.77 & 0.36 & 0.65 & \underline{0.87} & 0.51 & 0.24 & 0.40 & 0.66 \\
SVR                & 0.24 & 0.27 & 0.42 & 0.57 & 0.28 & 0.14 & 0.25 & 0.53 \\
XGBoost            & 0.75 & 0.37 & 0.68 & \underline{0.87} & 0.49 & 0.18 & 0.44 & 0.66 \\
TabPFN             & \textbf{0.86} & \textbf{0.52} & \textbf{0.74} & \textbf{0.90} & \underline{0.64} & \underline{0.25} & \underline{0.52} & \underline{0.74} \\
CNN & 0.78 & 0.33 & 0.58 & 0.79 & \textbf{0.65} & \textbf{0.42} & \textbf{0.62} & \textbf{0.75} \\

\bottomrule
\end{tabular}

\vspace{1ex}
\hspace{1ex}
\raggedright
For each performance metric, the best result within each feature-set block is shown \\
\hspace{1ex}
in \textbf{bold} and the second-best result is \underline{underlined}. \\
\hspace{1ex}
\textbf{Abbreviations:} GA (Geometric Averaged),
GW (Geometric Windowed), STW \\
\hspace{1ex}
(Spectro-Temporal Windowed).
\end{table}

\begin{table}[htbp]
\caption{Comparison of raw video-based pretrained CNN models for particle temperature and velocity prediction.}
\label{tab:raw_video_results}
\centering
\footnotesize
\setlength{\tabcolsep}{4pt}

\begin{tabular}{lcccccccc}
\toprule
\multirow{2}{*}{Model} & \multicolumn{4}{c}{Temperature Prediction} & \multicolumn{4}{c}{Velocity Prediction} \\
\cmidrule(lr){2-5} \cmidrule(lr){6-9}
 & $R^2$ & PAM$_{2.5}$ & PAM$_5$ & PAM$_{10}$ & $R^2$ & PAM$_{2.5}$ & PAM$_5$ & PAM$_{10}$ \\
\midrule
CNN+RH & \textbf{0.90} & \textbf{0.75} & \textbf{0.88} & \textbf{0.95} & \textbf{0.82} & \textbf{0.60} & \textbf{0.73} & \textbf{0.88} \\
CNN+LSTM & {0.75} & {0.63} & {0.77} & {0.84} & {0.81} & {0.49} & {0.70} & {0.87} \\
\bottomrule
\end{tabular}

\vspace{1ex}
\hspace{1ex}
\raggedright
For each performance metric, the best result is shown in \textbf{bold}. \\
\hspace{1ex}
\textbf{Abbreviations:} RH (Regression Head).
\end{table}

%\textbf{Abbreviations:} CNN (Convolutional Neural Network), LSTM (Long Short-Term \\
%\hspace{1ex}
%Memory), 

\paragraph{First Phase.} Table~\ref{tab:baseline_results} compares the primary modeling approaches across engineered feature representations. As shown in Table~\ref{tab:baseline_results}, TabPFN achieves the highest performance for particle temperature prediction when using the combined feature set (STW+GW+GA), reaching an $R^2$ of 0.86. In contrast, the CNN model achieves the best performance for particle velocity prediction across all four feature sets considered in the table. In particular, CNN attains its strongest velocity prediction performance when using the GW representation, with an $R^2$ of 0.81. The combined feature representation also performs well, suggesting that CNN models benefit from richer feature inputs that capture complementary aspects of plume dynamics. Results for the CNN model using the GA representation are not reported, as this configuration produced very poor performance. This behavior is likely due to the fact that GA features are temporally averaged summaries that remove the local temporal structure required by convolutional filters to capture meaningful patterns across adjacent temporal windows.

\paragraph{Second Phase.} Table~\ref{tab:raw_video_results} presents the performance of models that operate directly on raw video frames using pretrained convolutional feature extractors. In this setting, both evaluated architectures achieve strong predictive performance for particle temperature and velocity. Notably, the frozen pretrained CNN followed by a regression head (RH) achieves the strongest overall results, with $R^2=0.90$ for temperature prediction and $R^2=0.82$ for velocity prediction. This suggests that spatial representations extracted from plume imagery already contain substantial information related to particle characteristics. The CNN+LSTM architecture also achieves competitive performance,  although its performance remains slightly below that of the simpler CNN+RH model. 

We also evaluated an alternative architecture in which both the CNN architecture and LSTM layers were trained end-to-end on raw video frames. However, this configuration produced significantly worse performance. This behavior is likely due to the relatively small dataset size available for APS experiments. 

\paragraph{Statistical Uncertainty.} To quantify the uncertainty of the reported results, we compute 95\% confidence intervals for $R^2$ using a cluster bootstrap over spray runs~\cite{efron1993bootstrap}, in which $B=1000$ bootstrap resamples are drawn at the level of entire runs (each contributing its pre-spray and post-spray samples) with
replacement from the pooled out-of-fold predictions. For the best configurations, the intervals are: TabPFN (combined),
$R^2=0.86\,[0.68, 0.96]$ (temperature) and $0.64\,[0.45, 0.76]$ (velocity); CNN (GW), $R^2=0.81\,[0.66, 0.89]$ (velocity); and CNN+RH (raw video), $R^2=0.90\,[0.77, 0.98]$ (temperature) and $0.82\,[0.73, 0.91]$ (velocity). The wider temperature intervals reflect a few spray runs accounting for a disproportionate share of the error; even the lower bounds support strong predictive signal for both targets.

\section{Discussion}
\label{sec:discussion}

Across the engineered feature experiments, TabPFN consistently achieves the strongest or near-strongest performance for particle temperature, despite requiring no task-specific training or hyperparameter optimization unlike the other models. Although it cannot be applied directly to raw video, it is highly effective once structured feature
representations are constructed, making the feature engineering pipeline especially valuable for exploiting tabular foundation models in this low-data setting.

The engineered feature results also highlight the importance of representation design: combining STW, GW, and GA consistently outperforms the individual sets, indicating that geometric and spectro-temporal descriptors capture complementary aspects of plume behavior. CNN models on the engineered features perform strongly for velocity prediction,
reflecting the suitability of convolutional filters for capturing local cross-window interactions in the windowed representation.

The raw video experiments further show that direct visual representations of the plasma plume contain strong predictive information for particle characteristics. Both pretrained CNN-based architectures achieve strong performance for temperature and velocity prediction, indicating that spatial plume structure captured from video frames is highly informative. 
%Interestingly, the simpler frozen pretrained CNN followed by a regression head outperforms the CNN+LSTM architecture. This suggests that, in the current dataset, pretrained spatial representations already capture much of the relevant signal, while the limited data size makes more complex temporal modeling harder to exploit effectively.

From a deployment perspective, the study is a prototype-level
evaluation rather than an in-production demonstration, but several results support practical viability. The best-performing model places 95\% of temperature predictions and 88\% of velocity predictions within $\pm$10\% of the diagnostic ground truth (PAM@10\%, Table~\ref{tab:raw_video_results}), tolerance bands that are practically meaningful for maintaining coating consistency. Moreover, the proposed approach requires only a single fixed RGB camera, a lower-cost, less intrusive modality than commercial diagnostic systems. Since inference involves only a forward pass through lightweight models, the approach is computationally compatible with online operation. Validating these models under live production conditions remains future work.

A further consideration is generalizability across APS configurations. The present dataset, while spanning five gas-flow conditions and three nozzle diameters, was collected on a single torch with a fixed camera viewpoint and one feedstock material. Because the engineered descriptors and pretrained features both characterize plume appearance and dynamics, they are in principle transferable to other APS systems, though absolute intensity and geometry-based features are sensitive to camera placement, exposure, and
torch hardware, and would likely require recalibration before transfer. Robustness to longer-term process drift, such as electrode ageing over the torch lifetime, is also not directly assessed here, as our experiments used new electrodes throughout. Establishing cross-setup generalization would require multi-torch and multi-material data collection with domain-adaptation strategies, which
we leave for future work.

%Engineered features enable efficient use of tabular models such as TabPFN, while raw video models demonstrate that strong predictive signals are also present directly in plume imagery.

\section{Conclusion}
\label{sec:conclusion}

This work demonstrates that video-derived plume information can reliably predict key particle characteristics in APS. Using engineered video feature representations, TabPFN achieves the strongest and most consistent performance for particle temperature, while CNN models perform particularly well for particle velocity. 

%The results further show that combining complementary feature representations yields the most robust overall predictive performance.

Pretrained CNN-based models operating directly on raw video frames also achieve strong performance for both targets, showing that plume imagery itself contains highly informative signals for APS diagnostics.

%In particular, the frozen pretrained CNN followed by a regression head performs strongly, while the CNN+LSTM model provides a useful temporal baseline but does not outperform the simpler pretrained CNN model in the current data-limited setting.

These findings show that accurate APS particle prediction is achievable with small datasets given suitable visual representations, supporting video-based monitoring as a promising non-invasive approach for scalable, data-efficient diagnostics in thermal spray manufacturing.

Future work will explore multimodal fusion with additional sensing modalities, such as acoustic signals and process parameters, to further improve predictive accuracy and generalization across diverse APS operating conditions.

% Of course, authors have complete freedom on how they choose to structure their paper. Section headers from Introduction up to and including Conclusions are completely up to the discretion of the authors; use whichever structure you see fit. Title, Abstract, the credits environment, and References, however, are mandatory.

\begin{credits}

% Acknowledgements are hidden for double blind review
\subsubsection{\ackname}
The authors would like to thank McGill University, the National Research Council of Canada (NRC) as well as the National Program Office (NPO) for their support of this research. In particular, we acknowledge the technical staff at NRC Boucherville for their assistance with the atmospheric plasma spray experiments, data acquisition, and operation of the APS diagnostic systems.

\subsubsection{Data and Code Availability.}
The APS video recordings and particle diagnostic measurements used in this study were collected under a research collaboration with the National Research Council Canada and cannot be publicly released due to the confidentiality terms of the partnership. The feature extraction and modeling code is available from the authors upon reasonable request.

\subsubsection{\discintname}
The authors declare that they have no known competing financial interests or personal relationships that could have appeared to influence the work reported in this paper.

\subsubsection{Declaration of Generative AI in the writing process.}
During the preparation of this work, the authors used generative AI tools to assist with formatting, language refinement, and readability improvements. After using these tools, the authors carefully reviewed and edited the content and take full responsibility for the accuracy and integrity of the final content.

\end{credits}
\bibliographystyle{splncs04}
\bibliography{bibliography}
%% Note that this preceding line implies that you store your BibTeX references in a file called 'mybibliography.bib'. If you instead store your references in a file with a different name, for instance 'references.bib', the preceding line should read '\bibliography{references}'. Whatever you do, DO NOT put the file name extension .bib inside the \bibliography command; this will trip up LaTeX compilers. 
%
% If you do not want to use BibTeX, you can also type up the bibliography exactly as you see fit, using the following structure:
% % Note that this number 8 reserves an amount of space (equal to the natural width of the given number) for the label of your references; if you have more than 9 references, you will want to change this number to 18. If you have more than 19 references, this number is best changed to 88. If you have more than 99 references, I salute you.




% LNCS Homepage, \url{http://www.springer.com/lncs}, last accessed 2023/10/25
% \end{thebibliography}
\end{document}